%% file: main.tex
\documentclass{article}

\usepackage{hyperref}

\usepackage{microtype}
\usepackage{graphicx}
\usepackage{mathrsfs}
\usepackage{subfigure}
\usepackage{booktabs} 
\usepackage{tabularx}

\usepackage[accepted]{icml2024}

\usepackage{amsmath}
\usepackage{amssymb}
\usepackage{mathtools}
\usepackage{amsthm}
\usepackage{bbm}

\usepackage{algorithm}
\usepackage{algpseudocode}
\usepackage{tikz}
\usetikzlibrary{arrows.meta,calc}

\usepackage[capitalize,noabbrev]{cleveref}

\theoremstyle{plain}

\theoremstyle{definition}

\theoremstyle{remark}

\usepackage[textsize=tiny]{todonotes}

\icmltitlerunning{From Shortcut Learning to Discrete Neural Insertion Sort}

\begin{document}

\twocolumn[
\icmltitle{From Shortcut Learning to Discrete Neural Insertion Sort}


\icmlsetsymbol{equal}{*}

\begin{icmlauthorlist}
\icmlauthor{Konstantinos Mylonas}{}
\icmlauthor{Thrasyvoulos Spyropoulos}{}
\end{icmlauthorlist}

\icmlkeywords{Machine Learning, ICML}

\vskip 0.3in
]

\begin{abstract}
Neural algorithmic reasoning aims to train neural networks to follow known algorithms and generalize beyond the input sizes seen during training. However, correct final outputs and intermediate supervision do not necessarily show that a model follows the intended execution. We study this problem using insertion sort. Our analysis of the CLRS30 baseline NAR shows that the hint objective is weakly optimized and that hint accuracy remains low. Moreover, many intermediate representations can already be decoded into sorted sequences before the reference insertion-sort execution terminates, suggesting that the model learns a shortcut to the final output. Motivated by these findings, we introduce Discrete Neural Insertion Sort. Our model represents the sequence as a chain, separates scalar exchanges from control-state transitions, and projects node representations back to discrete states after every processor step. When trained only on sequences of length 16, the model achieves $100\%$ sorted-sequence accuracy on sequences of length 64 and 128. However, an ablation shows that discretization and graph structure alone are insufficient: without additional supervision of the global inner-loop state, the model fails even at the training length. Our results show that discrete execution can support strong length generalization, while also highlighting the problem-specific inductive bias required to learn a faithful algorithmic execution.
\end{abstract}


\input{01_introduction}
\input{08_related_work}
\input{02_background}

\input{03_diagnosing_trajectory_misalignment}
\input{04_discrete_neural_insertion_sort}

\input{05_experiements}
\input{06_limitations}
\input{07_conclusion}



\bibliography{example_paper}
\bibliographystyle{icml2024}

\end{document}

%% file: 01_introduction.tex
\section{Introduction}
\label{sec:introduction}

Neural networks have achieved impressive results in areas such as computer
vision, language generation, and reasoning. However, strong performance within
the training distribution does not always translate to reliable
out-of-distribution generalization \cite{geirhos2020shortcut}. This limitation can appear even in simple
algorithmic tasks. For example, neural networks trained to add real numbers
may fit their training distribution accurately while failing to recover the
exact addition rule or extrapolate outside that distribution
~\cite{klindt2023controlling}.

This behavior is particularly interesting because many of these tasks already
have known algorithmic solutions. The same addition procedure can be applied
to numbers of different magnitudes, and the same sorting algorithm can be
applied to sequences of different lengths. In contrast, a neural network may
learn statistical patterns that are specific to the examples observed during
training. Its predictions can therefore break down when the input size or
distribution changes.

Neural algorithmic reasoning (NAR) studies whether neural networks can instead
be trained to follow the steps of a known algorithm
~\cite{velivckovic2021neural}. Rather than learning only a direct mapping from
inputs to outputs, the model is encouraged to reproduce the intermediate
execution of the reference algorithm. If it learns the underlying transition
rule, the same neural processor may be applied repeatedly to inputs and
trajectories that are larger than those seen during training.

Our goal is not simply to train a neural network that can sort a sequence:
insertion sort already solves this problem exactly. Instead, we use insertion
sort as a controlled setting for studying whether a neural model can learn a
computation that is faithful to a reference algorithm, interpretable through
its intermediate states, and able to generalize beyond its training
distribution. More broadly, we ask what architectural constraints and
supervision are needed for such a computation to emerge through
gradient-based optimization.

The CLRS benchmark provides a common framework for studying this question
~\cite{velivckovic2022clrs}. In addition to the input and final output, it
provides \emph{hints} that describe the intermediate states of the reference
algorithm. A neural processor is supervised to predict these hints at every
execution step. The goal is to encourage the model to learn the transition
rules of the algorithm rather than a direct mapping from the input to the final
answer. If the learned transition is independent of the training size, the
same processor can be applied for more steps and on larger inputs.



However, accurate final predictions and intermediate supervision do not necessarily show that the model follows the intended algorithm. This distinction is especially important for sorting. Different sorting procedures produce the same final ordering while following very different intermediate trajectories. A model may therefore learn a shortcut that supports the final prediction without reproducing the execution that generated its hints.

In this work, we study this issue using insertion sort. We train the official CLRS baseline on insertion sort and examine both its hint predictions and its intermediate node representations. We find that the output objective is optimized much more effectively than the hint objective, while the accuracy of a representative hint remains low throughout the trajectory. We then apply the final-output decoder to embeddings produced at intermediate processor steps. For many inputs, these embeddings already decode to a sorted sequence several steps before the reference insertion-sort execution terminates. We refer to this behavior as \emph{rush-to-output}: the model learns representations that support the final answer without faithfully following the intended intermediate trajectory.

Motivated by this observation, we introduce \emph{Discrete Neural Insertion Sort}, a neural executor that more closely follows the control flow of insertion sort. The input sequence is represented as a bidirectional chain graph, reflecting the local comparisons and exchanges performed by the algorithm. Each node carries a scalar value and a discrete state describing its current role, such as $i$, $j$, or $k$ (see Fig.\ref{fig:insertion-sort-inner-loop} for an explanation of these roles). Scalar movement and discrete state transitions are handled separately. After every processor step, the updated node representations are projected back to discrete states, preventing arbitrary continuous node embeddings from being carried across the complete execution.

When trained only on sequences of length 16, the resulting model achieves
$100\%$ accuracy on
sequences of length 64 and 128. These results show that strong length
generalization is possible when the model learns the correct discrete
transition rule. However, our ablation shows that discretization alone is not
sufficient. Additional inductive bias is needed to guide the optimization
process toward the correct solution. This raises a broader question for neural
algorithmic reasoning: how much knowledge about an algorithm must be encoded
in the architecture and training objective for reliable execution to emerge?

Our contributions are as follows:

\begin{itemize}
    \item We analyze the CLRS30 insertion-sort baseline and show that its intermediate representations do not faithfully follow the supervised insertion-sort trajectory.
    \item We propose a discrete neural executor for insertion sort that separates scalar updates from control-state updates and restricts the recurrent computation through a discrete state bottleneck.
    \item We show that a model trained on sequences of length 16 can achieve perfect execution on sequences of length 64 and 128.
    \item We show that graph structure and discretization is not enough and additional inductive bias is required for this result.
\end{itemize}

%% file: 08_related_work.tex
\section{Related Work}
\label{sec:related-work}

\paragraph{Algorithmic alignment and length generalization.}
The principle of \emph{algorithmic alignment} suggests that a neural architecture is more likely to learn a target computation when its internal operations resemble the steps of that computation~\cite{Xu2019WhatCN}. Early work studied this idea using graph neural networks (GNNs), whose local message-passing operations naturally match the structure of many graph algorithms. ~\cite{pmlr-v336-nerem26a} further investigates this connection and provides conditions under which a carefully trained GNN can recover the Bellman--Ford execution and extrapolate to graphs outside its training distribution.

Neural algorithmic reasoning (NAR) develops this idea into a broader framework for training neural networks to execute algorithms~\cite{velivckovic2021neural}. One motivation for NAR is length generalization. Related work has also explored length generalization outside the standard NAR setting. For example, \cite{pmlr-v267-lee25d} trains a Transformer to add numbers with a small number of digits and then repeatedly retrains it using its own predictions on increasingly larger inputs, eventually extending the model to the desired number of digits~\

\begin{figure*}[t]
    \centering
    \includegraphics[width=\textwidth]{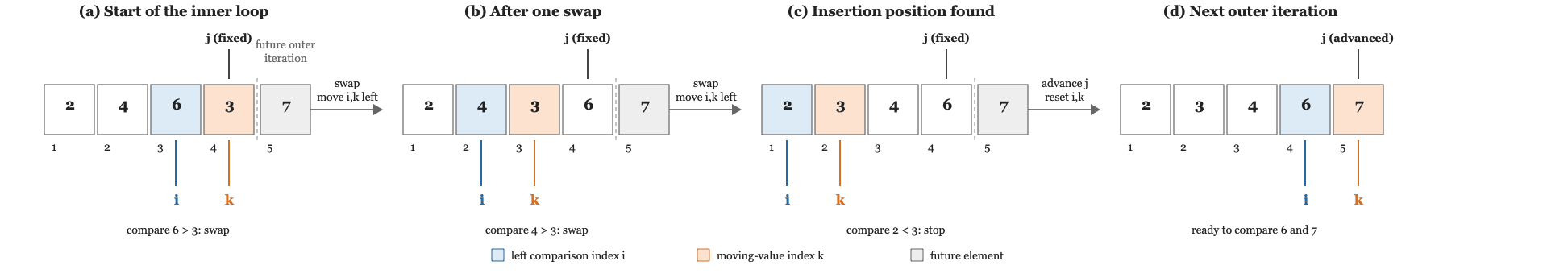}
    \caption{One inner-loop execution of insertion sort. The outer-loop index $j$ remains fixed at position 4 while $k$ tracks
        the value being inserted and $i$ identifies the element immediately to
        its left. Each successful comparison swaps the adjacent values and
        moves $i$ and $k$ one position to the left. The final comparison does
        not require a swap, so the inner loop terminates. The shaded node at
        position 5 has not yet been processed in panels (a)--(c). Panel (d)
        advances $j$ to this node and initializes $k=j$ and $i=j-1$, leaving
        the executor ready to begin the next inner loop.}
    \label{fig:insertion-sort-inner-loop}
\end{figure*}

\paragraph{The CLRS framework.}
The CLRS-30 benchmark established a common framework for neural algorithmic reasoning~\cite{velivckovic2022clrs}. It represents an algorithmic execution through its inputs, final outputs, and intermediate hints, and uses an encode--process--decode architecture to predict the execution one step at a time. ~\cite{ibarz2022generalist} improved the out-of-distribution performance of the original CLRS baseline and investigated whether a single shared processor could learn to execute many different algorithms. ForgetNet ~\cite{bohde2024markov} instead exploits the Markov structure of algorithmic execution by preventing the processor from directly carrying continuous hidden embeddings from one step to the next.


\paragraph{Faithfulness of neural execution.}
Most NAR work evaluates whether the model produces correct outputs on larger or shifted inputs. Comparatively less work has studied whether the intermediate neural computation faithfully follows the reference algorithm. ~\cite{pmlr-v231-mirjanic24a} analyzes the latent trajectories learned for Bellman--Ford by projecting them into two- and three-dimensional spaces and examining their structure. \cite{saldyt2025mindgapquantifyingmechanistic} studies faithfulness more directly for Bellman--Ford and BFS, comparing conventionally trained neural reasoners with models produced through neural compilation.

Our work contributes to this line of research by examining the execution learned by the CLRS insertion-sort baseline and provides evidence that a correct final result does not necessarily imply faithful algorithmic execution.

\paragraph{Discrete neural algorithmic reasoning.}
Discrete Neural Algorithmic Reasoning (DNAR) departs from the standard CLRS framework by introducing discrete node states and separating their updates from operations on continuous scalar values~\cite{pmlr-v267-rodionov25a}. DNAR achieves perfect out-of-distribution performance on several \textbf{graph algorithms}. The separation of discrete control and continuous values is also motivated by known difficulties in learning exact numerical operations. Neural Execution Engines ~\cite{NEURIPS2020_c8b9abff} identifies the representation and manipulation of numerical values as one source of difficulty in neural algorithmic execution. ~\cite{klindt2023controlling} demonstrates that models trained to add real numbers can fit the training distribution while failing to recover the exact operation or extrapolate beyond that distribution.
Our method is strongly motivated by DNAR but adapts its main ideas to insertion sort, a sequential algorithm rather than a graph algorithm. We represent the sequence as a chain, separate scalar exchanges from discrete control-state transitions, and introduce a virtual node to communicate global messages. In addition to demonstrating length generalization, we study the supervision and problem-specific inductive bias required for the model to learn the correct execution.

%% file: 02_background.tex

\section{Background}
\label{sec:background}


\subsection{Insertion Sort}
\label{sec:background-insertion-sort}

Insertion sort constructs a sorted prefix from left to right. At outer-loop
iteration $j$, the prefix $(s_1,\ldots,s_{j-1})$ is already sorted, and the
value initially stored at position $j$ must be inserted into its correct
position within this prefix. 

We distinguish three indices. The outer-loop index $j$ remains fixed throughout
one execution of the inner loop and marks the boundary of the prefix currently
being extended. The index $k$ tracks the value being inserted as it moves to
the left. The index $i=k-1$ identifies the value immediately to its left. If
$s_i>s_k$, the two values are exchanged and both $i$ and $k$ move one position
to the left. If $s_i\leq s_k$, the insertion position has been found and the
inner loop terminates. At the next outer-loop iteration, $j$ advances by one,
$k$ is reset to $j$, and $i$ is reset to $j-1$. 


Figure~\ref{fig:insertion-sort-inner-loop} illustrates one complete inner loop
for the sequence $(2,4,6,3,7)$. At the beginning of the illustrated outer-loop
iteration, $j=k=4$ and $i=3$. The value $3$ is exchanged first with $6$ and
then with $4$, while $j$ remains fixed. The comparison $2<3$ then identifies
the insertion position and terminates the inner loop. The node containing $7$
is retained in the figure to make clear that this is an intermediate
outer-loop iteration. Panel (d) shows the initialization of the following
iteration, with $j=k=5$ and $i=4$.

Two properties of this formulation are particularly relevant to our model.
First, the execution is sequential: the number of inner-loop steps depends on
the input values, and the next comparison is determined by the outcome of the
current one. Second, every scalar modification is local. A step either
exchanges the values at the neighboring positions $i$ and $k$ or leaves them
unchanged. These properties motivate our chain representation and the discrete
control states introduced in Section~\ref{sec:discrete-insertion-sort}.

\subsection{Neural Algorithmic Reasoning}
\label{sec:background-nar}

Neural algorithmic reasoning (NAR) aims to train a neural network to execute the steps of an algorithm~\cite{velivckovic2021neural}. The problem instance is first represented as a graph, and a neural processor, usually a graph neural network (GNN), updates this graph over a sequence of steps. The same processor is applied repeatedly, following the execution steps of the reference algorithm.

Algorithms usually require several steps before they terminate. To encourage the model to follow the correct execution, the training data contain information about the intermediate state of the reference algorithm. These intermediate signals are called \emph{hints}. At each processor step, the model predicts the hints for the next step. The idea is that supervising these predictions should encourage the model to learn the transition rule of the algorithm, rather than only a direct mapping from the input to the final output.

This framework was formalized by the CLRS benchmark through an \emph{encode--process--decode} architecture~\cite{velivckovic2022clrs}. Each example contains three types of information: \emph{inputs}, which describe the initial problem instance; \emph{hints}, which describe intermediate states of the algorithm; and \emph{outputs}, which describe its final result. We use insertion sort as a running example in this subsection. We refer the reader to the original CLRS benchmark paper for a more complete description of the framework.

For insertion sort, the input sequence is represented as a graph with one node for each sequence element. The processor operates on a fully connected graph, allowing every node to exchange messages with every other node. Each node is associated with two fixed input values: the scalar value of the corresponding sequence element and its position in the original sequence.

The CLRS representation also contains hints that change during the execution. One such hint is the predecessor pointer $\mathrm{pred}_h$, which describes the current ordering of the elements. These pointers are updated as insertion sort progresses and eventually describe the sorted order. The current indices $i$ and $j$ are also provided as hints. Together, these hints describe the current state of the reference insertion-sort execution.

At processor step $t$, encoders combine the input information and the current hints to produce latent node and edge embeddings. For example, the embedding of the node currently selected by $j$ should contain information indicating that this node has the role of $j$. The encoded graph is then passed to the processor GNN, where nodes exchange messages. A message from node $u$ to node $v$ can be written as

\begin{equation}
    \mathbf{m}_{u\rightarrow v}^{(t)}
    =
    M\left(
        \mathbf{z}_u^{(t)},
        \mathbf{z}_v^{(t)},
        \mathbf{e}_{uv}^{(t)}
    \right),
    \label{eq:nar-message}
\end{equation}

where $\mathbf{z}_u^{(t)}$ and $\mathbf{z}_v^{(t)}$ are the current embeddings of the two nodes, while $\mathbf{e}_{uv}^{(t)}$ contains the encoded information associated with their edge. Each node aggregates its incoming messages and uses them to update its embedding:

\begin{align}
    \overline{\mathbf{m}}_v^{(t)}
    &=
    \operatorname{AGG}_{u\in\mathcal{N}(v)}
    \mathbf{m}_{u\rightarrow v}^{(t)}, \\
    \mathbf{z}_v^{(t+1)}
    &=
    U\left(
        \mathbf{z}_v^{(t)},
        \overline{\mathbf{m}}_v^{(t)}
    \right).
    \label{eq:nar-update}
\end{align}

The updated embeddings should capture the changes that occurred during the current algorithmic step. Decoders map these embeddings back to the next hints. For example, suppose that the index $j$ moves from node $A$ to node $B$. After message passing, the updated embedding of node $B$ should contain enough information for the hint decoder to predict that $j$ is now located at node $B$. The predicted hints are then used in the following processor step, and the same processor parameters are reused throughout the execution.

Training commonly supervises both the intermediate hint predictions and the final output:

\begin{equation}
    \mathcal{L}_{\mathrm{NAR}}
    =
    \mathcal{L}_{\mathrm{output}}
    +
    \sum_t
    \mathcal{L}_{\mathrm{hint}}^{(t)}.
    \label{eq:nar-objective}
\end{equation}

The output loss measures whether the final answer is correct, while the hint losses measure whether the predicted intermediate states agree with the reference execution. During training, the ground-truth hints can be supplied at each step. During autoregressive execution, the model instead continues from its own predictions.

If the model learns the correct transition rule, the same processor can be applied for more steps than were observed during training. This provides a possible path to length generalization: a model trained on small graphs could execute the same local rule on larger graphs and over longer trajectories.

Hint supervision also makes it possible to study whether the model follows the intended algorithm. This is particularly useful for sorting because every correct sorting algorithm produces the same final ordering, even though the intermediate steps may be very different. A correct final output alone therefore does not show that the model executed insertion sort. The intermediate hints provide additional evidence about the trajectory followed by the model. As we show in the next section, however, supervising these hints does not guarantee that the learned continuous representations will follow the reference execution.

%% file: 03_diagnosing_trajectory_misalignment.tex
\section{Diagnosing Trajectory Misalignment in the CLRS Insertion-Sort Baseline}
\label{sec:baseline-diagnosis}

The motivation for hint supervision is that matching the intermediate states
of an algorithm should encourage a neural processor to reproduce its execution
trajectory. A low final-output error is insufficient evidence for this claim:
for a sorting task, a model can produce a sorted sequence without implementing
the particular sorting algorithm used to generate its supervision. We
therefore investigate whether the standard CLRS baseline follows the
insertion-sort trajectory encoded by its hints.

\subsection{Baseline and Diagnostic Instrumentation}
\label{sec:baseline-instrumentation}

We obtained the official CLRS implementation released by Google
DeepMind\footnote{\url{https://github.com/google-deepmind/clrs}}, trained it on insertion sort and extended it
with diagnostic instrumentation for insertion sort. The extensions serve two
purposes. First, we record the output and hint components of the training
objective separately. This reveals whether optimization of the final answer is
accompanied by optimization of the supervised intermediate trajectory. Second,
we apply the trained output decoder to the node embeddings produced at every
intermediate processor step. The standard model uses this decoder at the end of
the rollout; applying it earlier allows us to inspect when the latent
representations already contain the information required to predict a sorted
output.

The diagnostics in this section use sequences of length 16. The early-solution
experiment evaluates 64 sequences and compares the baseline's
intermediate decodes with the corresponding ground-truth insertion-sort
trajectories. These experiments evaluate the computation represented by the
model's intermediate states; they do not modify the definition of the CLRS
insertion-sort task.

\subsection{The Hint Objective Is Weakly Optimized}
\label{sec:baseline-hint-loss}

\begin{figure}[t]
    \centering
    \includegraphics[width=\columnwidth]{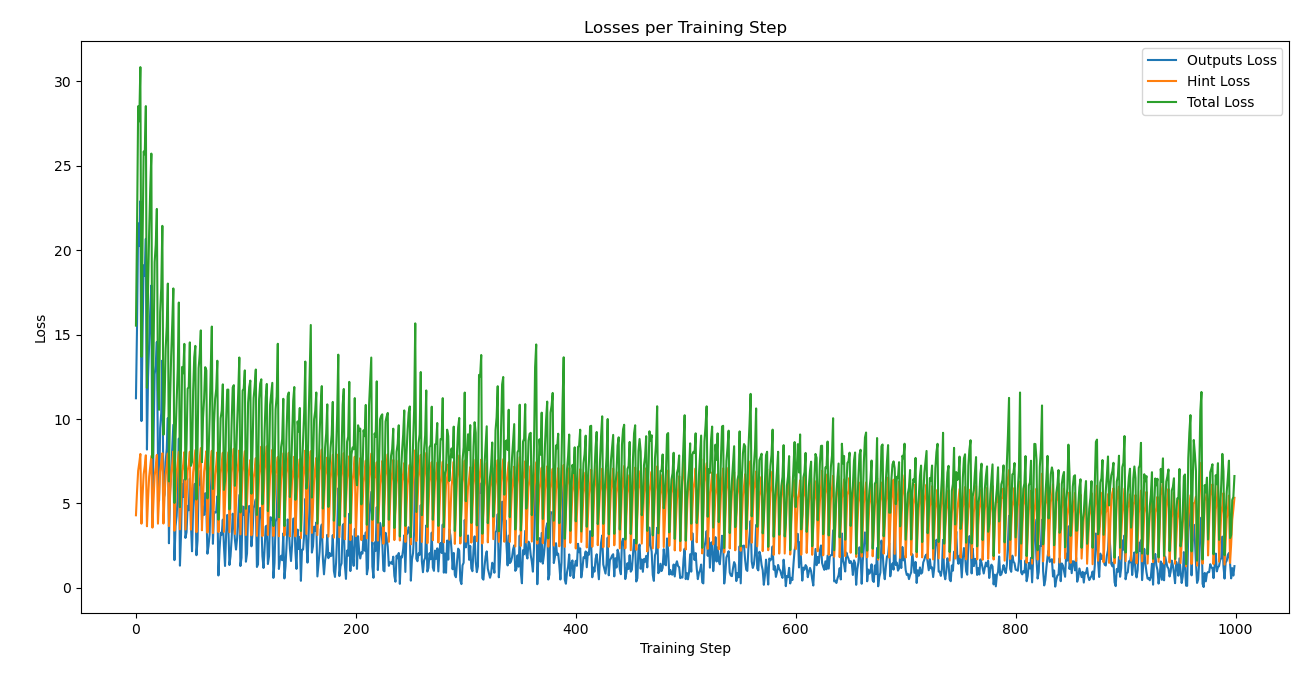}
    \caption{
        Output, hint, and total training losses for the CLRS insertion-sort
        baseline over 1,000 optimization steps. The output loss decreases
        substantially, whereas the hint loss remains high and strongly
        oscillatory. The hint component therefore does not exhibit the same
        sustained optimization as the final-output objective.
    }
    \label{fig:clrs-loss-components}
\end{figure}

Figure~\ref{fig:clrs-loss-components} separates the total training objective
into its output and hint components over 1,000 optimization steps. The output
loss decreases rapidly and remains comparatively low. The hint loss, however,
stays almost constant. Although it
exhibits a decreasing trend to some extent, it does not show the same sustained
optimization as the output objective.

\begin{figure}[t]
  \centering
  \includegraphics[width=\columnwidth]{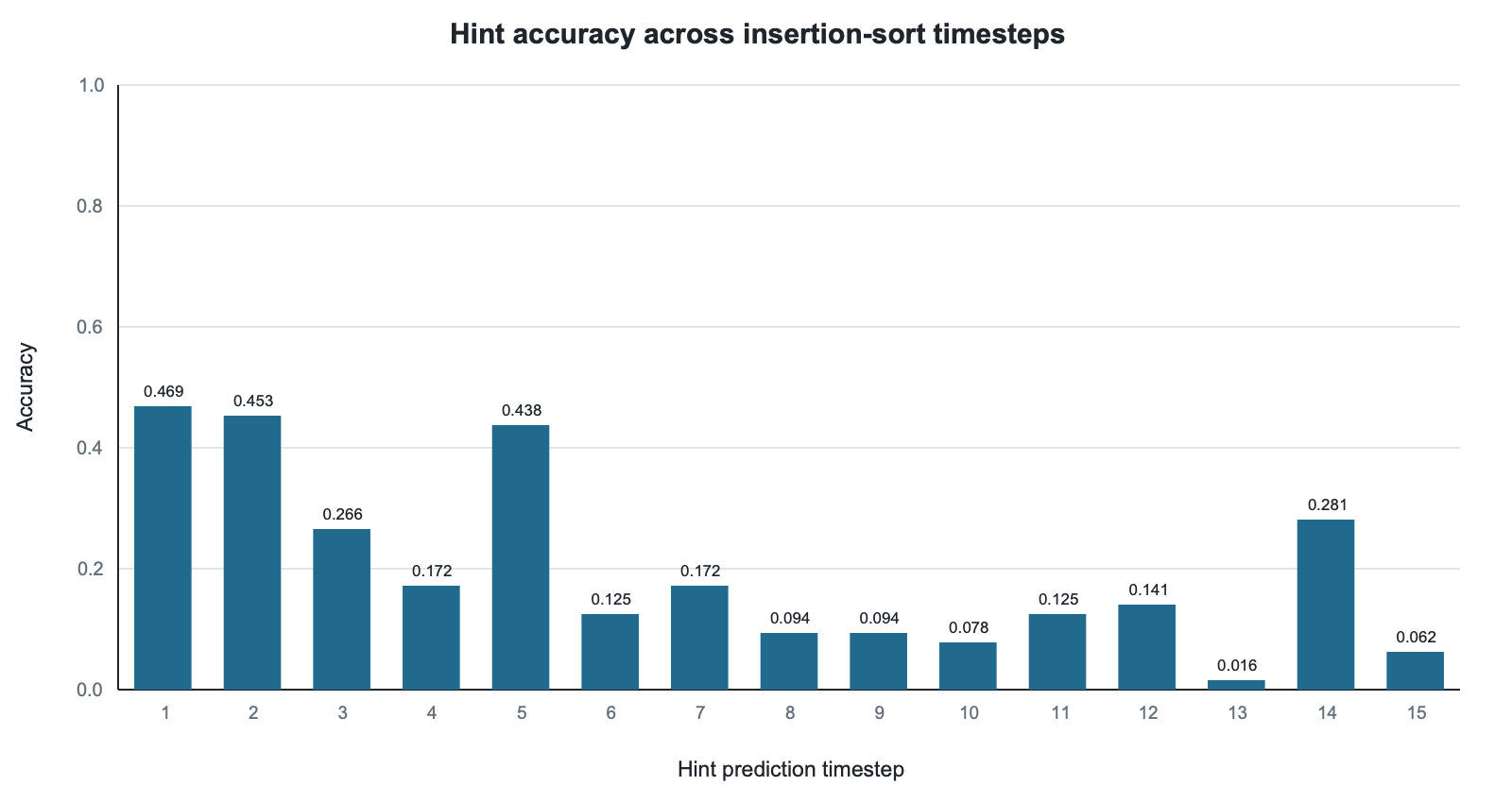}
  \caption{Average accuracy of a representative hint prediction at each of
  the 15 predicted transitions in a length-16 insertion-sort trajectory. The
  low accuracy provides further evidence that the model does not reliably
  follow the intermediate hint supervision.}
  \label{fig:hint-accuracy-by-timestep}
\end{figure}

In addition to the training loss, we record the average accuracy of a
representative hint prediction at each timestep. As shown in
Figure~\ref{fig:hint-accuracy-by-timestep}, the hint accuracy remains low across
the trajectory. This agrees with our intuition that the model is largely
ignoring the intermediate hint supervision while optimizing the final output.

The loss and accuracy measurements suggest that success on the final objective
does not require accurate reconstruction of the intermediate states. We
investigate this intuition more directly with our next diagnostic, which
probes the embeddings produced throughout the execution.

\subsection{Early Solutions}
\label{sec:early-solutions}

Let $T_x$ denote the step at which the \textbf{reference} insertion-sort trajectory for
an input $x$ reaches its sorted state. For an intermediate step $t<T_x$, we
take the baseline node embeddings produced at step $t$ and pass them through
the final-output decoder, treating them as if they were terminal embeddings.
We call the result an \emph{early solution} when this decoded output is already
a fully sorted sequence. This intervention does not change the model's
rollout. It asks when the information needed by the final decoder becomes
available in the latent trajectory.

\begin{figure*}[t]
    \centering
    \includegraphics[width=.8\textwidth]{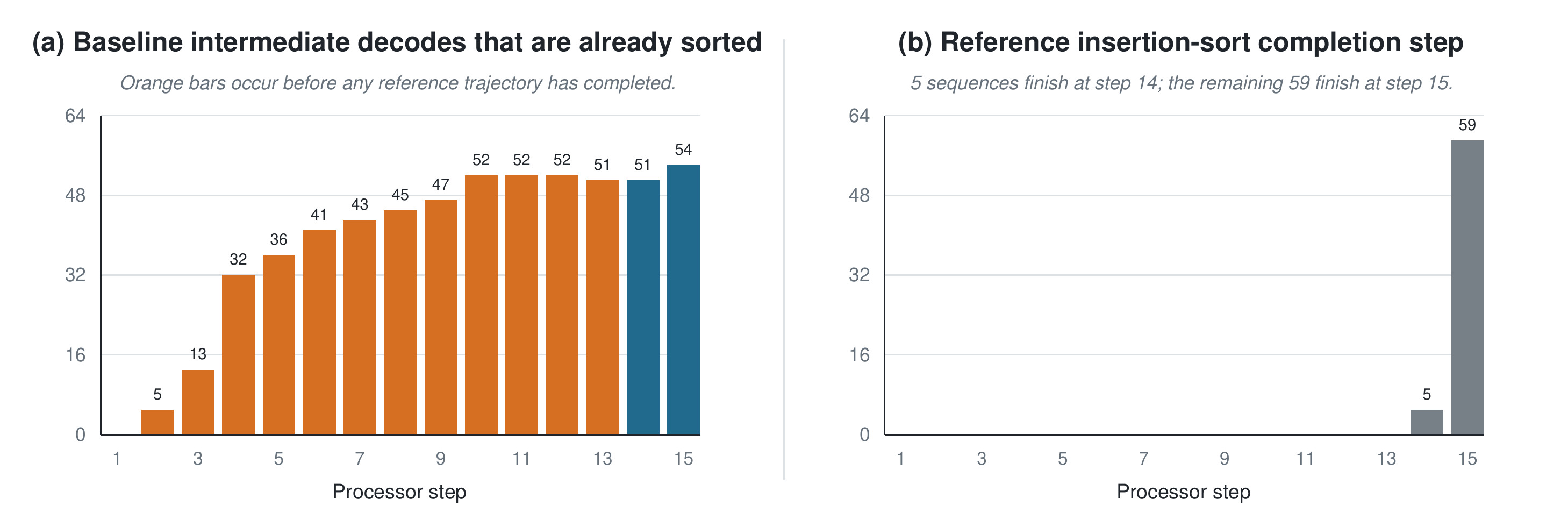}
    \caption{
        Early-solution diagnostic on 64 sequences of length 16. Panel (a)
        reports, at each intermediate processor step, how many baseline
        representations decode to a fully sorted sequence when passed through
        the final-output decoder. Panel (b) reports the step at which the
        corresponding reference insertion-sort trajectories actually finish.
        Thirteen baseline decodes are already sorted at step 3 and 32 are
        sorted at step 4, although none of the reference trajectories finish
        before step 14. Reference completion is concentrated at the final two
        steps: 5 sequences finish at step 14 and 59 at step 15.
    }
    \label{fig:early-solutions}
\end{figure*}

Figure~\ref{fig:early-solutions} shows the result. At processor step 3, the
intermediate embeddings already decode to sorted outputs for 13 of the 64
sequences, following 5 early solutions at step 2. At step 4, this number rises
to 32, or half of the evaluation batch.
By step 10, 52 sequences decode to sorted outputs. In contrast, none of the
reference insertion-sort trajectories finishes before step 14: 5 sequences
finish at step 14 and the remaining 59 finish at step 15. 



The discrepancy is substantial. For many inputs, the embeddings become useful
to the final-output decoder long before the supervised insertion-sort
trajectory reaches its terminal state. At the same time these intermediate embeddings are not particularly useful to the hint decoders as we see in Figure~\ref{fig:hint-accuracy-by-timestep}. The processor is therefore not only diverging from the correct hint sequence. It develops terminally useful
information earlier than it should if it followed the reference algorithmic trajectory.

\subsection{Shortcut Learning and Rush-to-Output Behavior}
\label{sec:rush-to-output}

Taken together, the two diagnostics indicate a mismatch between output
prediction and algorithmic execution. The model reduces the output loss while
ignoring hint loss, and its intermediate embeddings frequently decode
to sorted sequences many steps before insertion sort should finish. We
interpret this as a form of shortcut learning: the flexible continuous
processor discovers representations that support the final prediction without
faithfully maintaining the supervised intermediate state
sequence~\cite{geirhos2020shortcut}.

We refer to this phenomenon as \emph{rush-to-output behavior}. The model acts
as though it is preparing the terminal representation as early as possible,
while the hint trajectory becomes a secondary constraint. An early solution
does not imply that the model explicitly terminates its rollout, nor does a
single early decode prove which alternative sorting procedure it implements.
The output decoder is being applied at steps where it is not normally used.
Nevertheless, the prevalence of early solutions, together with the weak hint
optimization, provides evidence against the claim that the baseline's latent
states closely track its reference insertion-sort execution.

This diagnosis motivates the model introduced in
Section~\ref{sec:discrete-insertion-sort}. This model is fundamentally different than the framework established by the CLRS30 benchmark and is inspired by the work on Discrete Neural Algorithmic Reasoning \cite{pmlr-v267-rodionov25a}.


%% file: 04_discrete_neural_insertion_sort.tex

\section{Discrete Neural Insertion Sort}
\label{sec:discrete-insertion-sort}

We propose a neural model that follows the local steps of insertion sort. Each node has a discrete state, which represents its role in the algorithm, and a scalar value, which is the value being sorted. The discrete states determine which nodes participate in the current step, while a separate learned mechanism controls how scalar values move between nodes. This design makes each transition easier to inspect and constrains the model to operate with predefined discrete states.

\subsection{Sequence Representation}
\label{sec:sequence-representation}

Let the input be a sequence of scalars
$\mathbf{s}^{(0)}=(s_1^{(0)},\ldots,s_n^{(0)})$. We represent the sequence as a
bidirectional chain graph $G=(V,E)$, with one node for each sequence position
and two directed edges between every pair of consecutive positions:
\begin{equation}
    E=\{(u,u+1),(u+1,u)\mid 1\leq u<n\}.
    \label{eq:chain-edges}
\end{equation}
Consequently, information and scalar values can move by at most one sequence
position during a processor step. This locality matches the adjacent exchanges
performed by our insertion-sort trajectory and constitutes an explicit
structural inductive bias. Interior nodes have two incoming and two outgoing
edges, whereas the two boundary nodes have one of each.

\subsubsection*{Node Representations}
At time $t$, every node $u$ carries a scalar $s_u^{(t)}$ and a four-bit
discrete state $\mathbf{x}_u^{(t)}$. We use six discrete states which are defined in Table~\ref{tab:discrete-states}. At each timestep every node is in one of the six states and their states are updated between steps, keeping them aligned with the reference insertion sort execution.

\begin{table}[t]
    \centering
    \caption{Discrete node states used by the insertion-sort executor.}
    \label{tab:discrete-states}
    \small
    \begin{tabularx}{\columnwidth}{@{}l c X@{}}
        \toprule
        State & Code & Meaning \\
        \midrule
        $i$ & $1000$ & The left element in the current inner-loop comparison. \\
        $j$ & $0100$ & The current outer-loop element. \\
        $\mathrm{next}\_j$ & $0010$ & The node that becomes $j$ after the current inner loop terminates. \\
        $k$ & $0001$ & The right element in the current inner-loop comparison. \\
        $j \land k$ & $0101$ & The outer-loop element at the first step of its inner loop, when the roles of $j$ and $k$ coincide. \\
        nothing & $0000$ & A node with no active control role at the current step. \\
        \bottomrule
    \end{tabularx}
\end{table}

The 4-bit state is converted to an integer
and used to index a learned embedding table
$\mathcal{H}\in\mathbb{R}^{16\times d}$:
\begin{equation}
    \mathbf{h}_u^{(t)}
    =\mathcal{H}\!\left[\operatorname{bin}(\mathbf{x}_u^{(t)})\right].
    \label{eq:state-embedding}
\end{equation}
The processor operates
on these embeddings. After every processor step, the resulting continuous
representations are projected back through the discrete bottleneck described
in Section~\ref{sec:state-discretization}.

\subsubsection*{Edge Representations}
\label{sec:edge-features}

Insertion sort requires comparisons between values, but learning a numerical
comparator from finite training data introduces a source of
extrapolation error. We therefore provide the result of each local comparison
explicitly. For every directed edge $(u,v)$, we define an edge comparison bit
\begin{equation}
    c_{uv}^{(t)}
    =\mathbbm{1}\!\left[s_u^{(t)}>s_v^{(t)}\right]
    \label{eq:comparison-bit}
\end{equation}
and construct the edge representation
\begin{equation}
    \mathbf{z}_{uv}^{(t)}
    =\left[
        \mathbf{h}_u^{(t)}
        \mathbin{\Vert}
        \mathbf{h}_v^{(t)}
        \mathbin{\Vert}
        c_{uv}^{(t)}
    \right].
    \label{eq:edge-representation}
\end{equation}
Thus, the model does not learn the comparison operation itself. It learns how a
comparison outcome should affect scalar movement and control-state transitions,
conditional on the discrete roles of the incident nodes. Direction is
preserved: $\mathbf{z}_{uv}^{(t)}$ identifies $u$ as the sender and $v$ as the
receiver.

\subsection{Scalar Updates}
\label{sec:scalar-updates}

Our method separates scalar updates from discrete state updates as two completely independent processes. This section explains how the scalar values are updated between steps, while the following sections describe how the nodes update their discrete states.

Scalar values are updated using a learned gate, called \emph{take}, on each directed chain edge. For an edge from node $u$ to node $v$, a multilayer perceptron (MLP) produces the logit
\begin{equation}
a_{uv}^{(t)}
=
\operatorname{MLP}_{\mathrm{take}}
\left(\mathbf{z}_{uv}^{(t)}\right),
\label{eq:take-logit}
\end{equation}
where $\mathbf{z}_{uv}^{(t)}$ is the embedding of the directed edge $(u,v)$. During training, the gate is computed as
\begin{equation}
g_{uv}^{(t)}
=
\sigma\left(a_{uv}^{(t)}\right).
\label{eq:soft-take-gate}
\end{equation}
At inference time, we discretize it as $g_{uv}^{(t)} =
\mathbbm{1}\left[a_{uv}^{(t)}>0\right]$. The purpose of the \emph{take} gate is to decide whether the scalar value of the sender node should be transferred to the receiver node. This allows the model to perform a local swap between the nodes in states $i$ and $k$ when necessary. The scalar update is given by
\begin{equation}
s_v^{(t+1)}
=
\sum_{m\in\mathcal{N}(v)}
g_{mv}^{(t)}s_m^{(t)}
+
\left(
1-\sum_{m\in\mathcal{N}(v)}g_{mv}^{(t)}
\right)s_v^{(t)}
\label{eq:scalar-update}
\end{equation}

There are two possible cases in the reference algorithm.\footnote{We slightly abuse notation here. By $s_i$ and $s_k$, we mean the scalar values of the nodes whose discrete states are $i$ and $k$, respectively.} If $s_i>s_k$, the two nodes should swap their scalar values. If $s_i\leq s_k$, their values should remain unchanged. Therefore, when $s_i>s_k$, we want the gates on both directed edges between the $i$ and $k$ nodes to be active (i.e. $g_{ik} = g_{ki} = 1$. In all other cases, these gates should be inactive.

To see how Equation~\ref{eq:scalar-update} performs this operation, assume that the \emph{take} gate makes perfect predictions. Let node $v$ be in state $i$, node $u$ be in state $k$, and suppose that $s_v>s_u$. In this case, $g_{uv}=1$, while the gates from the other neighbors of $v$ are zero. Equation~\ref{eq:scalar-update} then becomes
\begin{equation}
s_v^{(t+1)} = s_u^{(t)}.
\end{equation}
Similarly, the gate on the reverse edge is active, so node $u$ takes the value of node $v$:
\begin{equation}
s_u^{(t+1)} = s_v^{(t)}.
\end{equation}
The two scalar values are therefore swapped. When no swap is needed, all incoming \emph{take} gates are zero, and Equation~\ref{eq:scalar-update} reduces to
\begin{equation}
s_v^{(t+1)}=s_v^{(t)}.
\end{equation}
We apply this update to every node at every execution step. The \emph{take} gate is trained using the mean squared error between the predicted and ground-truth scalar values. This error is backpropagated through the scalar update and into the MLP in Equation~\ref{eq:take-logit}.

\begin{figure}[t]
    \centering
    \includegraphics[width=.9\linewidth]{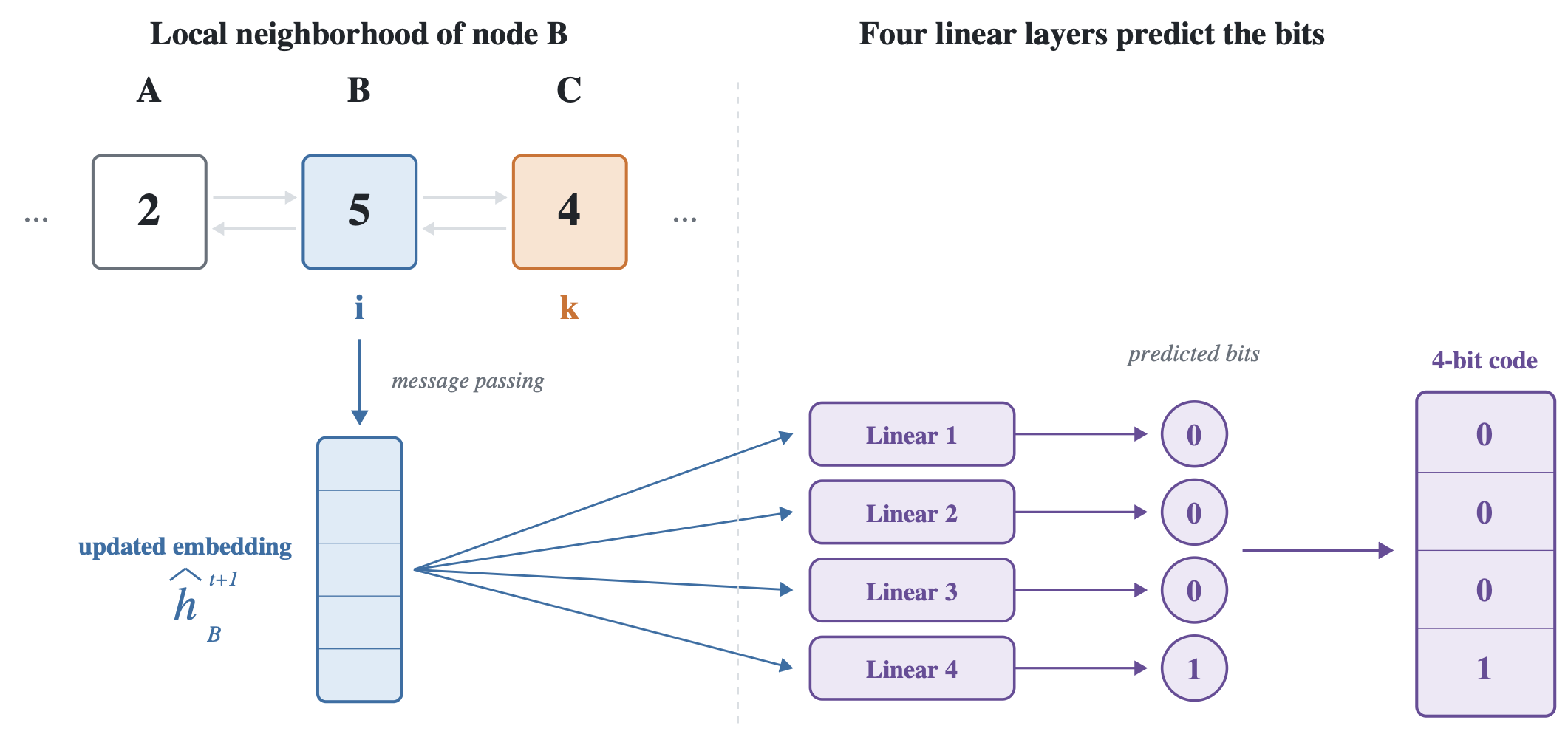}
    \caption{Discretization step: At timestep $t$, node $B$ is in state $i$. After message passing his updated embedding $\tilde{h}_B^{t+1}$ is passed through 4 linear layers to produce the new discrete state.}
    \label{fig:discretization-step}
\end{figure}

\begin{figure*}[t]
    \centering
    \includegraphics[width=.9\textwidth]{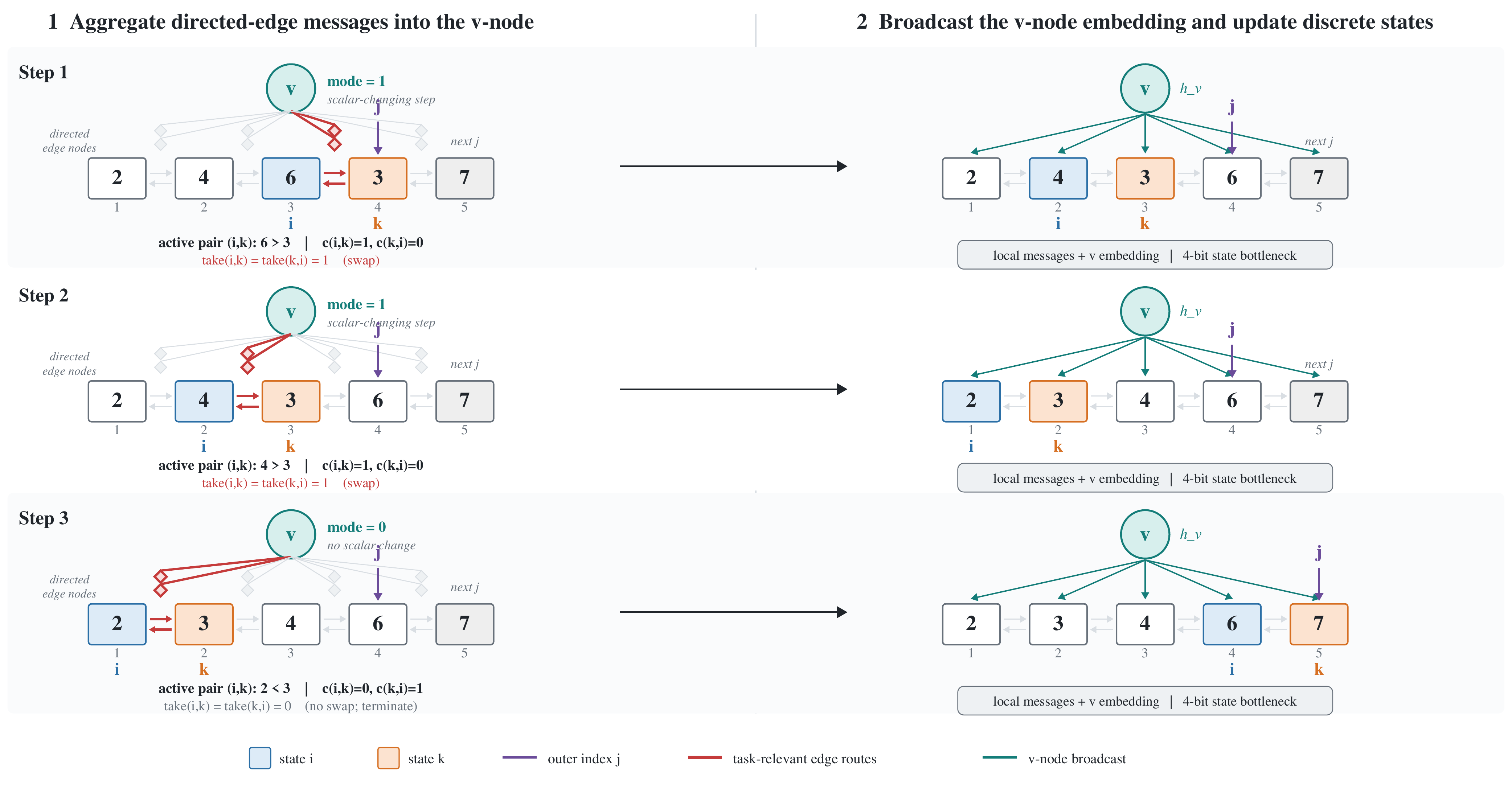}
    \caption{
        One inner-loop execution of the proposed discrete neural insertion-sort
        model. Each row is one processor step. In the first phase,
        the directed edges of the chain act as nodes in a virtual graph. The
        v-node aggregates their representations and must identify the two
        task-relevant directed edges between the active $i$ and $k$ nodes.
        The edge comparison bits $c$ help the v-node to decide if the inner loop is terminated and populate its embedding accordingly. In the second
        phase, the v-node broadcasts its embedding to every sequence node.
        Each node combines this graph-level context with local chain messages,
        and four binary state logits discretize the result. Steps 1 and 2 swap
        adjacent scalars and move $i$ and $k$ left. Step 3 produces no scalar
        change; the v-node communicates this global outcome, allowing the
        discrete transition to terminate the current inner loop and initialize
        the next one with $j=k=5$ and $i=4$. Faint routes are available to the
        model, while red routes mark the task-relevant edge messages the
        v-node is intended to identify.
    }
    \label{fig:method-overview}
\end{figure*}

\subsection{Discrete Updates}

In the previous section we saw how nodes swap their scalars whenever necessary in order to align with the reference insertion sort algorithm. This section focuses on how nodes update their discrete states.

\subsubsection*{Local Discrete Updates}
\label{sec:local-discrete-updates}

Section~\ref{sec:sequence-representation} introduced the six discrete states used by our method and explained how each state is mapped to a learned embedding. These embeddings are then used by the processor GNN. During each processor step, neighboring nodes exchange messages and use the received information to update their states.

For example, consider a node in state $i$ whose right neighbor is in state $k$. After receiving a message from the $k$ node, it can learn that it should take state $k$ at the next step (assuming that the inner loop continues). In this way, the discrete roles move along the chain as the execution of insertion sort progresses.

In standard neural algorithmic reasoning, the updated node embeddings are decoded to predict the next hints, while the continuous embeddings are passed directly to the following processor step. In our method, each updated embedding is instead projected back to a four-bit discrete state before the next step begins. This creates a discrete bottleneck between consecutive processor steps.

To perform this discretization, the updated embedding of each node is passed through four linear layers. Each layer predicts one bit of the node's next discrete state. Figure~\ref{fig:discretization-step} shows an example for node $B$, which is in state $i$ at step $t$. After message passing, its updated embedding $\widehat{\mathbf{h}}_B^{(t+1)}$ is passed through the four linear layers, producing the four-bit code of its discrete state at step $t+1$. In this example, the predicted code is $0001$, so node $B$ moves to state $k$.


\subsubsection*{Global Inner-Loop Context}
\label{sec:virtual-node}

The local chain structure lets the model perform an adjacent exchange, but some state transitions require distant information. For example the node in state $\mathrm{next}\_j$ can't decide when to switch to state $j$ only from his local neighbors' messages. The transition to $j$ happens only when the inner loop is terminated. A node cannot determine from local state alone whether the current inner-loop
step caused a scalar change elsewhere in the graph. We
introduce one virtual node $v_G$ per input graph to make this graph-level context available to every sequence node.

The virtual-node computation treats each directed edge of the original chain as
an item in a second, virtual graph. First, edge representations are embedded:
\begin{equation}
    \mathbf{r}_{uv}^{(t)}
    =\operatorname{MLP}_{\mathrm{edge}}
      \!\left(\mathbf{z}_{uv}^{(t)}\right).
    \label{eq:virtual-edge-embedding}
\end{equation}
A learned scoring layer assigns a score to each directed edge, and a softmax
over all edges in the input graph produces attention coefficients:
\begin{equation}
    \beta_{uv}^{(t)}
    =\frac{
        \exp\!\left(\mathbf{w}_{v}^{\top}\mathbf{r}_{uv}^{(t)}\right)
    }{
        \sum_{(p,q)\in E}
        \exp\!\left(\mathbf{w}_{v}^{\top}\mathbf{r}_{pq}^{(t)}\right)
    }.
    \label{eq:virtual-attention}
\end{equation}
The virtual-node embedding is then
\begin{equation}
    \mathbf{h}_{v_G}^{(t)}
    =\operatorname{MLP}_{v}\!\left(
        \sum_{(u,v)\in E}
        \beta_{uv}^{(t)}\mathbf{r}_{uv}^{(t)}
    \right).
    \label{eq:virtual-node-embedding}
\end{equation}
This mechanism allows the virtual node to focus on the edge between the active
$i$ and $k$ nodes and to summarize whether their scalar comparison produced a
swap. The virtual node has a directed outgoing edge to every sequence node, so
this summary can affect all state transitions during the same processor step.


\subsubsection*{Discrete Updates with Global Information}
\label{sec:state-discretization}

Having introduced the discrete state updates and the use of global context, we can now describe how the complete method works. Figure~\ref{fig:method-overview} illustrates the process through several execution steps. This figure hides the details of the scalar updates, as they are updated through an independent mechanism that we described in Section \ref{sec:scalar-updates}.

A processor step begins with the construction of the virtual node's embedding. The virtual node receives messages from all edges and learns to focus on the important ones through the attention mechanism. The virtual node's role is to identify when an inner loop is terminated. The required information to make this decision is available on the directed edges, as their representations are constructed based on the incident nodes' states and their scalar comparison bit $c$.

After computing the virtual-node embedding, we augment the original chain graph with
the edges $(v_G,u)$ for all $u\in V$. A single attention head performs message
passing on this augmented graph. For each edge $(p,u)$, including virtual-node
edges, the attention score is
\begin{equation}
    \ell_{pu}^{(t)}
    =\frac{
        (W_Q\mathbf{h}_u^{(t)})^{\top}
        (W_K\mathbf{h}_p^{(t)})
    }{\sqrt{d}},
    \label{eq:node-attention-score}
\end{equation}
and incoming messages are aggregated as
\begin{equation}
    \widetilde{\mathbf{h}}_u^{(t+1)}
    =\mathbf{h}_u^{(t)}
    +\sum_{p:(p,u)\in E'}
      \alpha_{pu}^{(t)}W_V\mathbf{h}_p^{(t)},
    \label{eq:node-message-passing}
\end{equation}
where $E'=E\cup\{(v_G,u):u\in V\}$. The coefficients $\alpha$ are normalized
over the incoming edges of each receiver.The implementation uses softmax attention.

The updated embeddings are discretized as we discussed earlier to produce the new discrete states. The resulting four-bit prediction is used to retrieve the discrete embedding
for the next recurrent step. This bottleneck prevents arbitrary continuous node
features from being carried across the execution trajectory.


\subsection{Training and Execution}
\label{sec:training-and-execution}

Our model receives supervision at multiple parts of each processor step. As is common in neural algorithmic reasoning, intermediate supervision is used to encourage the model to follow the trajectory of the reference algorithm rather than only predict the final output.

First, we supervise the discrete state updates. The four linear layers produce one logit for each bit of the new node state, and we apply binary cross entropy between the predicted and ground-truth bits. Second, we supervise the scalar updates using the mean squared error between the updated scalar values and the corresponding values from the reference insertion-sort execution.

Finally, we also supervise the virtual node. We attach a linear prediction head to the v-node embedding and train it to distinguish between steps in which the inner loop continues and steps in which it terminates. This auxiliary objective encourages the v-node embedding to capture the global information needed by the nodes to update their discrete states. The prediction head is used only to provide supervision during training and is not required during inference.

The complete loss for one processor step is therefore
\begin{equation}
    \mathcal{L}^{(t)}
    =
    \mathcal{L}_{\mathrm{state}}^{(t)}
    +
    \mathcal{L}_{\mathrm{scalar}}^{(t)}
    +
    \lambda_v \mathcal{L}_{v}^{(t)},
    \label{eq:per-step-loss}
\end{equation}
with $\lambda_v=1$ by default. Training is teacher forced: at every recorded
step, the processor receives the ground-truth node states and scalars and
predicts their values at the following step. This isolates the learning of the
local transition function from errors accumulated earlier in a trajectory.

Evaluation is autoregressive. Starting from the initial states and input
scalars, the executor repeatedly feeds its predicted discrete states and
updated scalars back into the same processor. Since the processor parameters do
not depend on the sequence length or execution step, the learned transition
rule can be applied for the longer trajectories required by sequences outside
the training-length distribution.

%% file: 05_experiements.tex
\section{Experiments and Ablations}
\subsection{Length Generalization}
\label{sec:length-generalization}

\begin{table}[t]
    \centering
    \caption{Performance across different sequence lengths. The model is trained only on sequences of length 16 and evaluated autoregressively without additional training or fine-tuning.}
    \label{tab:length-generalization}
    \small
    \begin{tabular}{lccc}
        \toprule
        \textbf{Sequence length}
        & \textbf{State}
        & \textbf{Scalar}
        & \textbf{Sorted} \\
        \midrule
        16  & $100\%$ & $100\%$ & $100\%$ \\
        64  & $100\%$ & $100\%$ & $100\%$ \\
        128 & $100\%$ & $100\%$ & $100\%$ \\
        \bottomrule
    \end{tabular}
\end{table}
We first evaluate whether our model can generalize to sequences that are longer than those used during training. Training examples are generated by sampling scalar values independently from a uniform distribution over $[0,1]$ and running the reference insertion-sort implementation to obtain the complete sequence of scalar values and discrete states. We train only on sequences of length 16. A new batch of 32 sequences is generated at every optimization step, and the model is trained for 1,000 epochs.

We report three evaluation metrics. \emph{State accuracy} is the percentage of node states for which all four predicted bits match the ground-truth discrete state. \emph{Scalar accuracy} is the percentage of scalars that end up in the correct position. Finally, \emph{sorted-sequence accuracy} is the percentage of input sequences that are sorted correctly. All metrics are measured during autoregressive execution: after the initial step, the model receives its own predicted states and scalar values rather than the ground-truth trajectory. Table~\ref{tab:length-generalization} summarizes our results on sequences of both the training length and larger, unseen lengths.

On sequences of the training length, the model reaches $100\%$ accuracy on all three metrics. We then evaluate the same trained model on sequences of length 64 and 128, without any additional training or fine-tuning. The model again achieves $100\%$ state accuracy, scalar accuracy, and sorted-sequence accuracy.

These results show that the learned execution rule transfers to trajectories much longer than those seen during training. Since the model is forced to pass through a discrete state after every processor step, it cannot carry arbitrary continuous information across the full execution. Instead, it must repeatedly apply the learned local transitions. When these transitions are learned correctly, the same model can execute insertion sort on much larger sequences.

Achieving this result, however, required enough supervision to guide the model toward the correct transitions. In the next subsection, we study the role of this supervision and show that the auxiliary objective on the virtual node is necessary for reliable length generalization.

\subsection{The Role of Virtual-Node Supervision}
\label{sec:inductive-bias}

The results in the previous section show that our model can generalize to sequences much longer than those used during training. However, this result should not be attributed to the discrete bottleneck alone. Our method includes several sources of prior knowledge about insertion sort, including the chain structure, the comparison bits on the edges, the predefined discrete states, and the intermediate supervision of state and scalar updates. This raises a natural question: how much of this inductive bias is needed for the model to learn an execution that generalizes?

Initially, we expected the supervision of the discrete states and scalar values to be sufficient. Since the virtual node is computed from the representations of all edges, the model has access to the information required to determine whether the current inner loop should continue or terminate. In principle, this information could be learned indirectly through the state and scalar objectives.
In practice, we found that this was not the case. Without direct supervision on the virtual node, the model failed to learn the correct execution rule, even for sequences of the training length. Although it received supervision on the discrete states and scalar updates, it did not learn a virtual-node embedding that reliably represents the state of the inner loop. We therefore introduced the auxiliary virtual-node objective described in Section~\ref{sec:training-and-execution}. This objective uses a linear head to predict whether the current inner-loop step continues or terminates.

Table~\ref{tab:v-node-ablation} reports the sorted-sequence accuracy with and without virtual-node supervision. Without this supervision, the model fails to sort the sequences correctly, including sequences of length 16. Once the auxiliary objective is introduced, the model reaches $100\%$ accuracy at the training length and maintains this performance on sequences of length 64 and 128. This result suggests that simply making the relevant information available to the model does not guarantee that it will use that information in the intended way. The auxiliary objective guides the virtual-node embedding toward the global information required by the discrete state transitions.

\begin{table}[t]
    \centering
    \caption{Effect of supervising the virtual node. Both variants use the
    same architecture and receive supervision on the discrete states and
    scalar updates. We report the sorted sequence accuracy.}
    \label{tab:v-node-ablation}
    \small
    \begin{tabular}{lccc}
        \toprule
        \textbf{Model}
        & \textbf{$n=16$}
        & \textbf{$n=64$}
        & \textbf{$n=128$} \\
        \midrule
        Without v-node supervision & $0\%$ & $0\%$ & $0\%$ \\
        With v-node supervision    & $100\%$ & $100\%$ & $100\%$ \\
        \bottomrule
    \end{tabular}
\end{table}

The ablation shows that successful length generalization depends not only on
restricting the recurrent state, but also on supervising the information that
the discrete transitions require. In our case, the difficult part was the
global decision that marks the end of an inner loop. Local state and scalar
supervision did not consistently cause this information to emerge in the
virtual-node embedding. A direct auxiliary objective was needed to make this
global signal reliable.

%% file: 06_limitations.tex
\section{Limitations and Future Work}
\label{sec:limitations}

Our method is designed specifically for insertion sort. We believe that its main ideas, could also be applied to other sequential algorithms. However, extending the method would require identifying an appropriate graph representation, a set of discrete states, and any global information needed by the algorithm. Extending this method to become more generic and applicable directly or with minimal effort to other sequential algorithms is therefore an interesting direction for future work.

Our method also differs from the standard neural algorithmic reasoning framework. In standard NAR models, the recurrent computation is performed using continuous latent representations. Our model still uses continuous embeddings during each message-passing step, but it does not allow these embeddings to be passed freely between consecutive execution steps. This restriction makes the execution easier to inspect, but it also reduces the freedom of the neural processor.

Another limitation is the amount of problem-specific knowledge included in the model. The chain graph encodes the local structure of insertion sort, the edge comparison bit provides the result of each scalar comparison, and the discrete states describe the control flow of the algorithm. Even with this structure and direct supervision of the state and scalar transitions, the model did not reliably learn the required global information without additional supervision on the virtual node. This suggests that providing the necessary information to the model is not always sufficient: the training objective must also guide the model toward using that information in the intended way.

The resulting generalization therefore comes with a tradeoff. Stronger supervision and more algorithm-specific structure make the learned execution more reliable, but they also require us to provide more knowledge about the solution in advance. An important direction for future work is to investigate whether the same behavior can be learned with weaker supervision, or whether the required discrete states and global signals can be discovered automatically.

%% file: 07_conclusion.tex
\section{Conclusion}
\label{sec:conclusion}

In this work, we studied whether neural algorithmic reasoning models faithfully execute insertion sort. Our experiments with the CLRS baseline showed that good final predictions do not necessarily mean that the model follows the reference algorithm. The hint objective remained weakly optimized, while intermediate embeddings often decode to a sorted solution ealry, before the reference insertion-sort execution had finished.

Motivated by these observations, we introduced a discrete neural executor for insertion sort. The model separates scalar values from discrete control states and restricts the information carried between processor steps through a discrete bottleneck. When trained on sequences of length 16, it achieves perfect autoregressive execution on sequences of length 64 and 128.

Our experiments also show that this generalization requires strong inductive bias and carefully chosen supervision. In particular, direct supervision of the virtual node was needed for the model to learn the global signal used by the discrete state transitions. These results suggest that discrete execution can support strong length generalization, while also showing how much knowledge about the target algorithm may need to be included in the model and its training objective.